\documentclass[fleqn,10pt]{wlscirep}
\usepackage[utf8]{inputenc}
\usepackage[T1]{fontenc}

\usepackage{subfiles}
\usepackage{multirow}
\usepackage{tikz} 

\usepackage{amsthm,amsmath,amssymb}
\usepackage{mathptmx}
\usepackage{mathrsfs}
\DeclareMathAlphabet{\mathcal}{OMS}{cmsy}{m}{n}  
\usepackage{ulem} 

\usepackage{subcaption}
\usepackage{tabularx}
\usepackage[table]{xcolor}
\definecolor{tableHead}{RGB}{216,214,194}
\definecolor{tableContent}{RGB}{235,234,222}

\definecolor{modification}{RGB}{0,0,0}  

\definecolor{cred}{HTML}{FF6B6B}
\definecolor{cyellow}{HTML}{FEC260}
\definecolor{cgreen}{HTML}{6BCB77}
\definecolor{cgreen}{HTML}{70AD47}
\definecolor{cblue}{HTML}{4D96FF}
\definecolor{cpurple}{HTML}{2A0944}
\definecolor{ggray}{RGB}{127,127,127}
\definecolor{aliceblue}{rgb}{0.94, 0.97, 1.0}

\usepackage{makecell}   

\newcommand{\modelname}{ChemCRAFT}

\title{
 Agentic reinforcement learning empowers next-generation chemical language models for molecular design and synthesis
}

\author[1,2,3*]{Hao Li}
\author[2*]{He Cao}
\author[1,3*]{Shenyao Peng}
\author[2]{Zijing Liu}
\author[2]{Bin Feng}
\author[1,3]{Yu Wang}
\author[1,3]{Zhiyuan Yan}
\author[1,3]{Yonghong Tian}
\author[2$\dagger$]{Yu Li}
\author[1,3$\dagger$]{Li Yuan}
\affil[1]{\textit{School of Electronic and Computer Engineering, Peking University Shenzhen Graduate School, Shenzhen, 518055, China}}
\affil[2]{\textit{International Digital Economy Academy~(IDEA)}}
\affil[3]{\textit{Peng Cheng Laboratory, Shenzhen, 518000, China}}

\affil[*]{These authors contributed equally to this work}
\affil[$\dagger$]{Corresponding authors: liyu@idea.edu.cn, yuanli-ece@pku.edu.cn
}

\begin{abstract}

Language models are revolutionizing the biochemistry domain, assisting scientists in drug design and chemical synthesis with high efficiency. Yet current approaches struggle between small language models prone to hallucination and limited knowledge retention, and large cloud-based language models plagued by privacy risks and high inference costs.
To bridge this gap, we introduce \modelname{}, a novel framework leveraging agentic reinforcement learning to decouple chemical reasoning from knowledge storage. Instead of forcing the model to memorize vast chemical data, our approach empowers the language model to interact with a sandbox for precise information retrieval. This externalization of knowledge allows a locally deployable small model to achieve superior performance with minimal inference costs.
To enable small language models for agent-calling ability, we build an agentic trajectory construction pipeline and a comprehensive chemical-agent sandbox. Based on sandbox interactions, we constructed ChemToolDataset, the first large-scale chemical tool trajectory dataset. Simultaneously, we propose SMILES-GRPO to build a dense chemical reward function, promoting the model's ability to call chemical agents.
Evaluations across diverse aspects of drug design show that \modelname{} outperforms current cloud-based LLMs in molecular structure analysis, molecular optimization, and synthesis pathway prediction, demonstrating that scientific reasoning is not solely an emergent ability of model scale, but a learnable policy of tool orchestration. 
This work establishes a cost-effective and privacy-preserving paradigm for AI-aided chemistry, opening new avenues for accelerating molecular discovery with locally deployable agents. Code available at \href{https://github.com/HowardLi1984/ChemCraft}{https://github.com/HowardLi1984/ChemCraft}.

\end{abstract}
\begin{document}
\flushbottom
\maketitle

\section*{Introduction}
Chemical Language Models (CLMs) have emerged as transformative tools in accelerating drug discovery and materials science~\cite{han2025generalist, zhang2024scientific, xia2023systematic,lv2025navigate}, demonstrating remarkable potential in tasks ranging from molecular property prediction~\cite{molformer, chemberta, chemberta2} and de novo design~\cite{grisoni2023chemical, bhattacharya2024large, bagal2021molgpt,lv2025prollama} to retrosynthesis planning~\cite{wan2022retroformer, liu2024multimodal, ma2025automated} and reaction condition recommendation~\cite{qian2023predictive, zhang2025large, cao-etal-2024-presto}. Currently, the development of CLMs primarily follows two distinct paradigms. The first approach involves supervised fine-tuning (SFT), or the continued pre-training of large language models on domain-specific corpora~\cite{taylor2022galactica, edwards2022translation, zhang2024chemllm, zhao2024chemdfm, xia2025nature}. While these models achieve competitive performance on in-distribution metrics~\cite{edwards2022translation, pei2023biot5, zhuang2024instructbiomol}, they fundamentally suffer from rigid optimization objectives. By forcing the model to minimize loss on "Task-Query-Direct Answer" triplets, current SFT paradigms compel the LLM to learn a shallow mapping from chemical inputs directly to numerical values or labels~\cite{Ding2024BreakTCA, li2025detect, Rueda2025UnderstandingLSA}. Critically, this approach bypasses the "expert-like" reasoning process—such as structural analysis, intermediate hypothesis generation, and logical verification—that is essential for scientific discovery~\cite{zhang2025exploring, truhn2023large, shojaee2024llm}. Consequently, this leads to a degradation of the model's intrinsic general capabilities, often referred to as catastrophic forgetting~\cite{luo2025empirical, liu2024more, zheng2025towards}, and limits its adaptability across disparate chemical tasks~\cite{toniato2023fast, zhao2023scientific, ganeeva2024lost}.

To bridge this cognitive gap, recent state-of-the-art works have introduced the concept of "Chemical Reasoning"~\cite{ouyang2023structured, jang2025structural, zhao2025chemdfm, bran2025mist, zhao2025molreasoner, li2025mol, zhuang2025reasoning, wang2025chem, narayanan2025training}. By utilizing distillation from superior teacher models~\cite{zhao2025molreasoner, zhuang2025reasoning, wang2025chem} and Reinforcement Learning (RL) based post-training, these approaches aim to guide models in mimicking the step-by-step, coherent workflows of human experts. 
While enhancing interpretability, this paradigm introduces a critical bottleneck. We find that the reasoning process is frequently dominated by token-intensive, low-level tasks—such as valency checks and structural parsing~\cite{alampara2025general, mswahili2024transformer}—rather than high-level derivation. This strategy is intrinsically inefficient. Given the probabilistic nature of LLMs, utilizing them for deterministic rote calculations is error-prone~\cite{bran2023chemcrow, m2024augmenting, mirza2025framework}. More importantly, it diverts the model's limited context window away from complex chemical reasoning toward mere syntactic validation~\cite{yao2022react, liu-etal-2024-lost}.

Alternatively, Multi-Agent Systems (MAS) have attempted to mitigate this by offloading tasks to external tools via commercial LLM APIs~\cite{bran2023chemcrow, m2024augmenting, boiko2023autonomous, song2025multiagent} (e.g., GPT-4). However, this introduces severe deployment bottlenecks: the prohibitive token costs for large-scale screening and the unacceptable privacy risks associated with transmitting proprietary molecular structures to cloud servers~\cite{feretzakis2024privacy, guo2025survey, muegge2024perspectives}. To resolve these challenges, we present \modelname{}, a next-generation chemical language model framework designed to democratize high-performance, privacy-preserving intelligence in chemical research. Moving beyond the "internalize-everything" dogma, we establish a "cognitive decoupling" architecture inspired by the human scientific workflow~\cite{kahneman2011thinking, evans2003two}. In this paradigm, \modelname{} functions as the central scientific reasoner, formulating hypotheses and orchestrating a comprehensive chemical sandbox of external tools. This synergy enables the model to solve complex problems within a joint language-tool space~\cite{schick2023toolformer, mialon2023augmented}, ensuring that verified tools maintain scientific rigor while the model's parameters are dedicated to high-level planning and logic.

To enable the training of such a system, we constructed and open-sourced a large-scale dataset of expert-level tool-use trajectories. Leveraging this data, our work challenges the prevailing assumption that complex tool orchestration is an emergent ability exclusive to massive, proprietary LLMs (>100B parameters)~\cite{wei2022emergent}. We demonstrate that a compact 7B-14B parameter model, when optimized via supervised fine-tuning and domain-specific reinforcement learning, can achieve tool-use capabilities comparable to commercial APIs. By aligning the model's policy with the "Hypothesis-Action-Observation"~\cite{yao2022react} scientific loop, we successfully elicit robust reasoning and self-correction in smaller architectures. This breakthrough effectively resolves the deployment trilemma of cost, performance, and privacy, enabling every laboratory to host a secure, expert-level AI chemistry copilot locally.

To rigorously validate the application potential of our framework across diverse chemical domains, we employ ChemCoTBench~\cite{li2026chemcotbench}, a comprehensive evaluation suite designed to assess the multi-step reasoning and tool-use capabilities of chemical agents. Covering a spectrum of nine critical tasks—ranging from fundamental molecular structure understanding and editing to molecular property optimization and reaction-related tasks—this benchmark provides a holistic view of an agent's proficiency. Extensive evaluations reveal that \modelname{} achieves a new state-of-the-art among open-source models. Remarkably, despite its compact parameter size, it demonstrates problem-solving capabilities that not only significantly surpass the baselines of similar scales but also rival, and in specific reasoning-intensive tasks, exceed those of leading commercial LLM APIs.

\section{Results}

\begin{figure*}    
\includegraphics[width=1.0\linewidth]{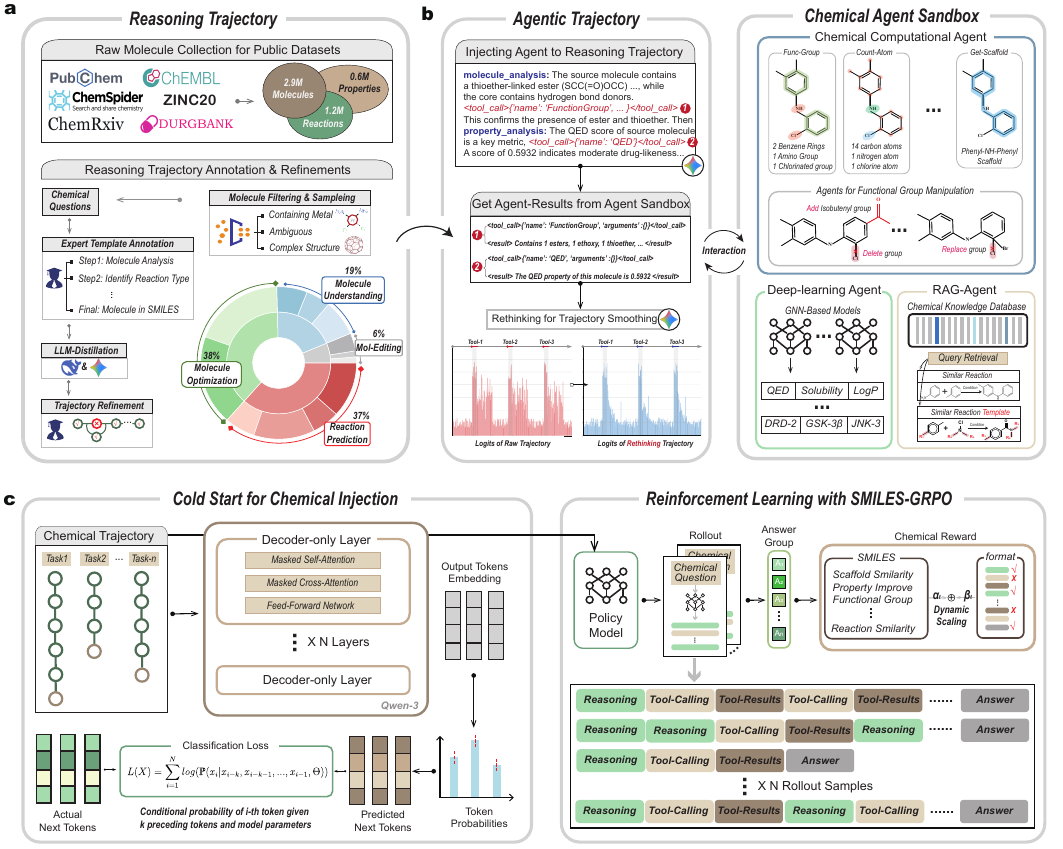}
    \caption{
    Overview of the data-curation pipeline and the training pipeline of our \modelname{}. \textbf{a} We build up the agentic trajectory dataset based on the pure reasoning trajectory with the interaction of our Chemical Agent Sandbox. \textbf{b} We implement a two-stage training scheme. The first stage involves cold-start training using token-level loss to enable the model to memorize the tool invocation format. The second stage employs GRPO to enhance the model’s understanding of tools.
    }
    \label{fig:framework}
    \vspace{-0.1in}
\end{figure*}

\subsection{Framework Construction}
To operationalize the "Cognitive Decoupling" paradigm, we established a systematic pipeline for data curation and model training, as illustrated in Figure~\ref{fig:framework}. This pipeline transforms raw chemical data into high-quality "Agentic Trajectories" and subsequently aligns the model’s capabilities through a two-stage training process involving cold-start SFT and domain-specific reinforcement learning.

\paragraph{Structuring the Chemical Space with ChemCoTBench.}
Our data curation begins with the aggregation of raw molecular structures and reaction data from public repositories, including PubChem, ChEMBL, and ZINC. However, simply feeding models with raw data is insufficient for cultivating scientific intelligence, as current evaluations often reduce chemistry to simple factual recall or property prediction, neglecting the step-by-step operational reasoning required for real-world discovery. To bridge this gap and provide a rigorous "anchor" for our learning objectives, we align our framework with ChemCoTBench. Unlike traditional benchmarks, ChemCoTBench systematically decomposes the chemical discovery process into 9 major tasks and 22 subtasks, bridging fundamental reasoning with high-stakes practical applications. Specifically, it connects foundational Molecule Understanding and Editing directly to downstream challenges: Molecule Optimization serves as a proxy for de novo drug design (e.g., optimizing binding affinity for targets like DRD2 or GSK3-$\beta$), while Reaction Prediction (including forward prediction and retrosynthesis) underpins automated chemical synthesis planning. By decomposing these complex challenges into explicit sequences of modular chemical operations (e.g., addition, deletion, substitution), ChemCoTBench provides the essential "scaffold" for our work, converting abstract chemical challenges into actionable, traceable reasoning steps that mirror the rigorous workflows of professional chemists.

\paragraph{Innovating Data Construction: Tool-Integrated Trajectories.} Traditionally, constructing reasoning datasets relies on distilling traces from powerful closed-source LLMs by prompting them to mimic expert thought patterns. However, this static textual approach fails to capture the interactive, hypothesis-driven nature of the modern scientific workflow. To transcend this limitation, we propose a novel Tool-Integrated Reasoning Trajectory construction method, as visualized in Figure~\ref{fig:framework}.

First, to ground the model's reasoning in computational reality, we engineered the Chemical Agent Sandbox—a high-performance execution environment where essential chemical computations (e.g., RDKit structure parsing, QED/LogP evaluation, and reaction database retrieval) are refactored into scalable microservices. This sandbox encapsulates domain complexity, allowing the LLM to drive professional-grade calculations via standardized API interfaces. This design effectively decouples reasoning from calculation, preventing the model from struggling with error-prone internal arithmetic and allowing it to focus on high-level strategy.

Building upon this infrastructure, we construct the training trajectories through a "Hypothesis-Action-Refinement" loop (Figure~\ref{fig:framework}a). A critical innovation here is our "Reflective Refinement" mechanism. Raw tool-use logs are often disjointed and mechanical (e.g., "Action: Calculate MW; Observation: 150"). Our mechanism transforms this by injecting the verified tool outputs back into the context and prompting the teacher model to rewrite its reasoning process. The model preserves the original intent and the tool's result, but weaves them into a fluid, logical analysis. This process effectively converts rigid API logs into dynamic, expert-level scientific narratives, where the agent interprets evidence, validates hypotheses, and adjusts its strategy in a manner consistent with human domain experts.

\paragraph{The Two-Stage Training Paradigm.}
Leveraging these curated trajectories, we employed a progressive training strategy to instill both chemical knowledge and agentic behavior. 
\textbf{Stage 1: Cold-Start SFT:} We first subjected the base model (7B/14B) to supervised fine-tuning using the synthesized agentic trajectories. This "Cold Start" phase is critical for initializing the model's understanding of chemical syntax and establishing the fundamental "Think $\rightarrow$ Call Tool $\rightarrow$ Observe" behavioral pattern, providing a stable policy foundation for subsequent optimization.
\textbf{Stage 2: Reinforcement Learning with Chemical-Aware Rewards:} To transcend mere imitation and achieve robust problem-solving, we advanced to a reinforcement learning framework utilizing Group Relative Policy Optimization (GRPO)~\cite{guo2025deepseek}. Departing from generic scalar feedback mechanisms, we engineered a multidimensional chemical reward function designed to evaluate the scientific validity of the entire reasoning chain; this signal rigorously integrates structural validity via exact SMILES matching and scaffold similarity, functional fidelity through the alignment of functional groups and reaction templates, and optimization success based on the magnitude of property improvement (e.g., $\Delta$LogP, $\Delta$QED). By optimizing against these granular chemical metrics rather than simple text overlap, the model learns to strategize its tool usage and reasoning path to maximize scientific accuracy.

\subsection{Chemical Proficiency Evaluation}
Authenticating an AI agent's utility in chemical research demands an evaluation that transcends rote memorization, probing its ability to navigate the full hierarchy of chemical complexity—from fundamental structural parsing to the intricate logic of synthesis planning. Utilizing the ChemCoTBench framework, we subjected our \modelname{} models to a rigorous assessment across fundamental molecular understanding, property-guided optimization, and reaction prediction, benchmarking them against premier reasoning-enhanced LLMs and domain-specific baselines. Our empirical analysis reveals that by offloading precise computation to a specialized sandbox, smaller agentic models can surmount the inherent trade-off between semantic reasoning and structural precision, achieving expert-level proficiency that rivals or eclipses proprietary frontier models in rigorous scientific tasks.

\begin{table*}[!htp]
\centering
\footnotesize
\begin{subtable}{\textwidth}
  \setlength{\tabcolsep}{4.0mm}        
  \centering
  \begin{tabular}{l|cc|cc|c|ccc}
    \toprule
    \multirow{2}{*}{Models} & \multicolumn{2}{c}{Func-Group} & \multicolumn{2}{c}{Scaffold} & \multicolumn{1}{c|}{SMILES} & \multicolumn{3}{c}{Molecule-Edit} \\
    \cmidrule(r){2-3} \cmidrule(r){4-5} \cmidrule(r){6-6} \cmidrule(r){7-9}
    & FG$\downarrow$ & Ring$\downarrow$ & Murcko$\uparrow$ & Ring-sys$\uparrow$ & Eq.$\uparrow$ & Add$\uparrow$ & Delete$\uparrow$ & Sub$\uparrow$\\
    \midrule
    Gemini-2.5-Pro-think & {0.11} & {0.60} & {0.51} & {87.5} & 82 & \textbf{100.0} & {85.0} & 81.7 \\
    Claude3.7-sonnet-think & 0.21 & 1.60 & 0.40 & 80.0 & {84} & 85.0 & 80.0 & \textbf{83.4}\\
    DeepSeek-R1 & 0.27 & 1.55 & 0.34 & 45.0 & 65 & 70.0 & 70.0 & 68.3\\
    O3-Mini & 0.13 & {0.60} & 0.39 & 75.0 & 78 & 65.0 & 55.0 & 80.0\\
    Deepseek-V3 & 0.15 & 1.50 & 0.24 & 76.7 & 77 & 70.0 & 75.0 & 76.7\\
    Qwen3-235B-A22B & 0.42 & 1.00 & 0.34 & 82.5 & 75 & 40.0 & 75.0 & {66.7}\\
    Qwen2.5-32B-Instruct & 0.36 & 0.65 & 0.12 & 53.3 & 62 & 50.0 & 50.0 & 48.3\\
    \midrule
    \modelname{} & \textbf{0.03} & \textbf{0.15} & \textbf{0.57} & \textbf{100} & \textbf{97} & 85.0 & \textbf{95.0} & 80.0 \\ 
    \bottomrule
  \end{tabular}
\end{subtable}
\vspace{0.2cm}

\begin{subtable}[t]{0.8\textwidth}
\footnotesize
\centering
\begin{tabular}{cccc}
\toprule
     \textbf{Function-Group-Detection} & \textbf{Ring-System-Detection} & \textbf{Molecule-Edit-Substitute}  \\
\midrule
    \begin{tabular}[b]{c}
    \includegraphics[width=0.35\textwidth]{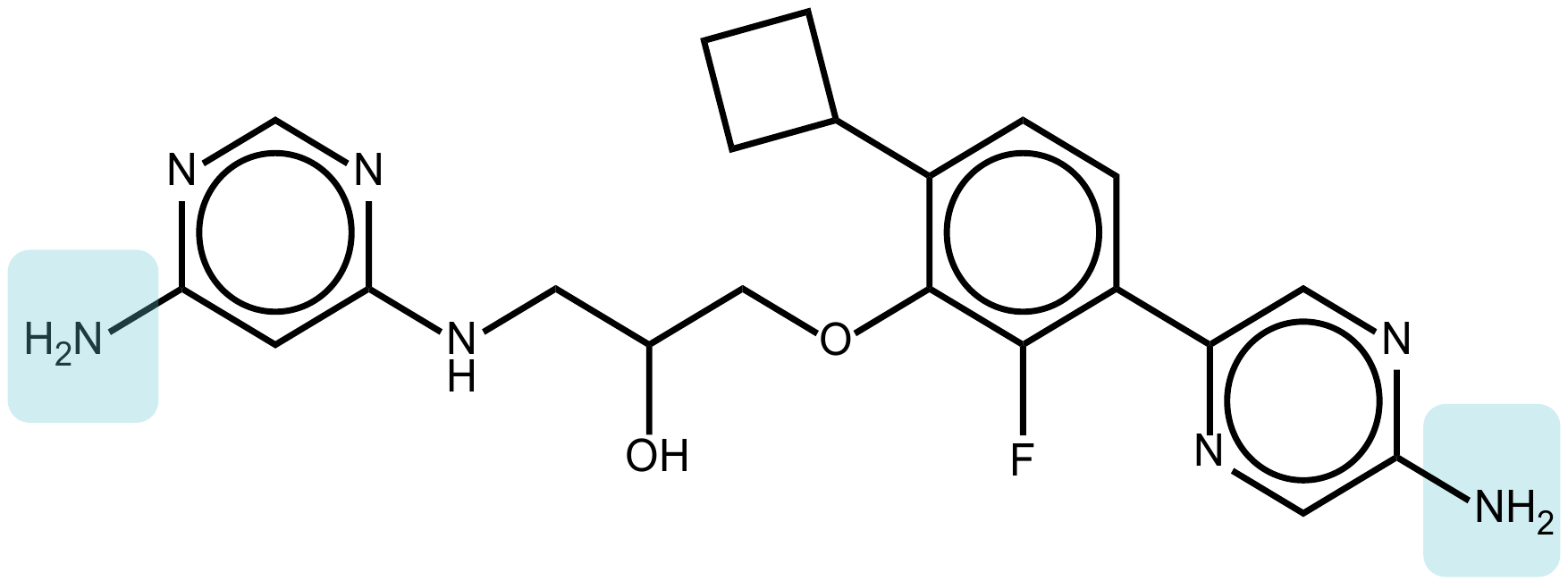} 
    \end{tabular} &
    \begin{tabular}[b]{c}
    \includegraphics[width=0.25\textwidth]{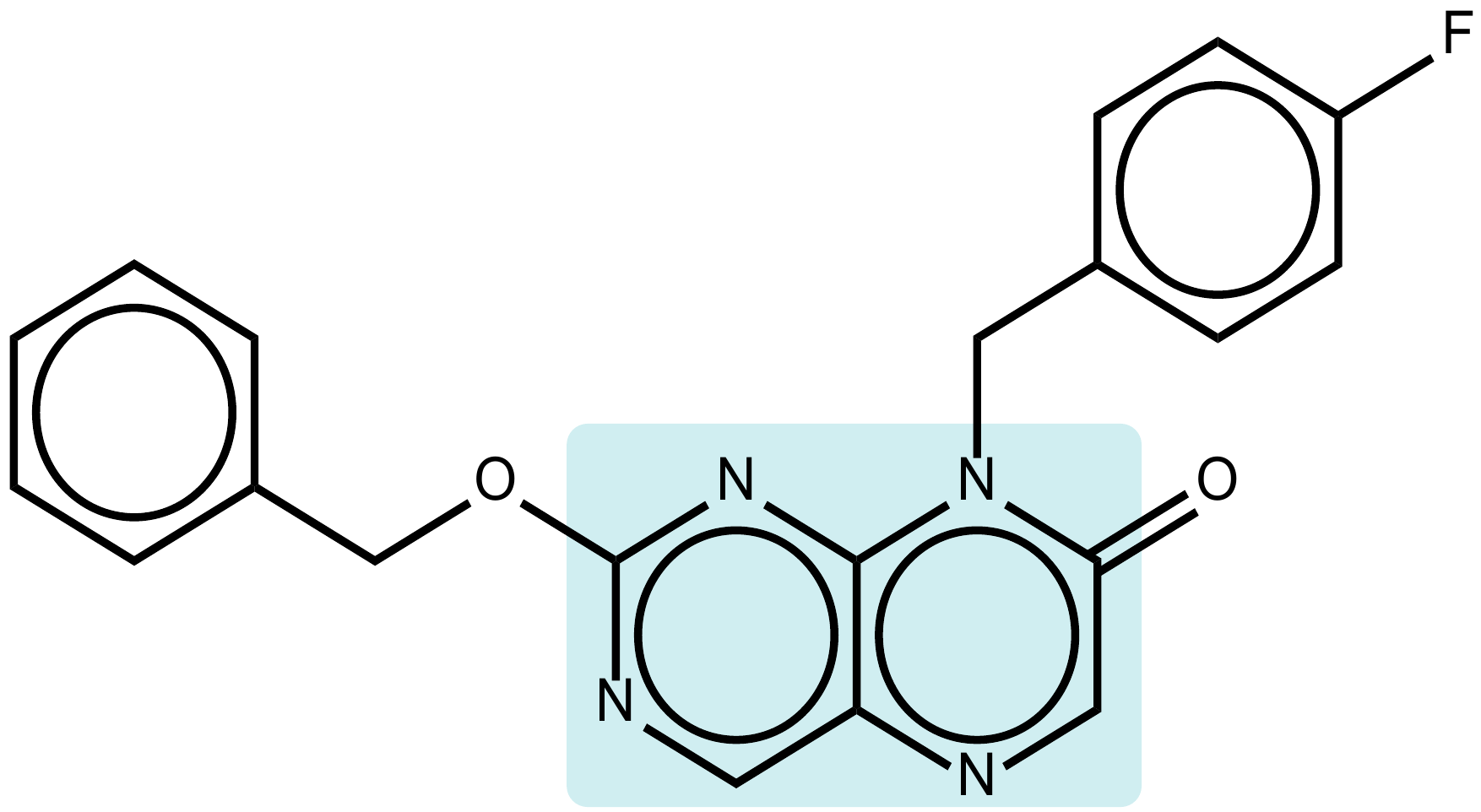} 
    \end{tabular}
    &
    \begin{tabular}[b]{c}
    \includegraphics[width=0.25\textwidth]{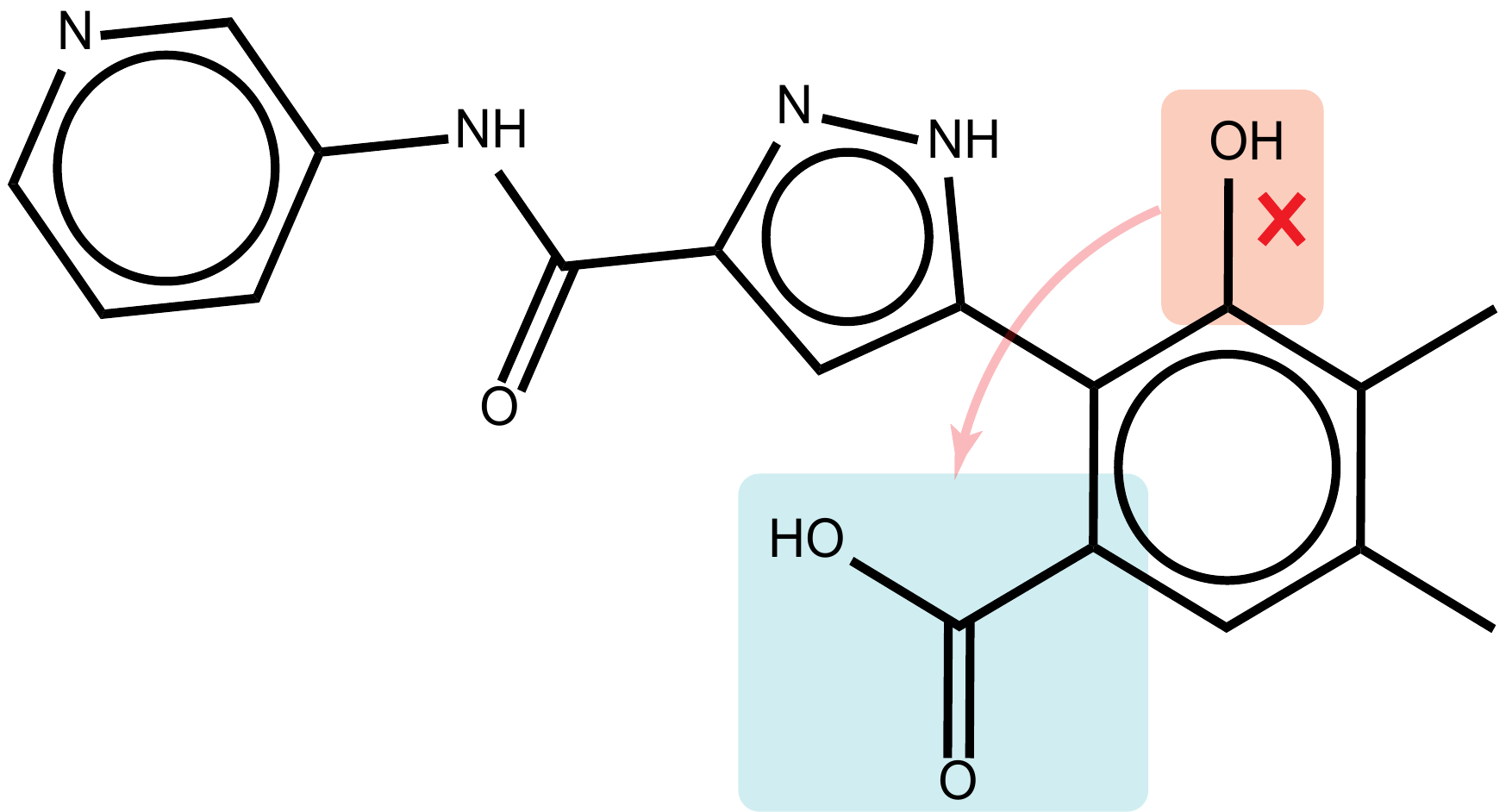} 
    \end{tabular} \\
    \begin{tabularx}{0.35\textwidth}{@{}X@{}}
    Giving you an input molecule, help me count the \textbf{amino group} number in the molecule.
    \end{tabularx}
    &
    \begin{tabularx}{0.25\textwidth}{@{}X@{}}
    Determine whether the ring system is in the target molecule. 
    \end{tabularx}
    & 
    \begin{tabularx}{0.25\textwidth}{@{}X@{}}
    Help me delete the \textbf{hydroxyl} group and add \textbf{carboxyl} group.
    \end{tabularx}
    \\
\bottomrule
\end{tabular}
\end{subtable}
\caption{\textbf{Performance Comparison for Molecule Understanding and Molecule Editing.} We propose the experimental results on our \modelname{} and baselines, including biochemical models and advanced LLMs.} 
\label{tab:molund_edit}
\end{table*}

\paragraph{Mastering Molecular Syntax and Logic.}
Fundamentally, an AI chemist must accurately perceive and manipulate molecular structures before it can engage in complex design. To rigorously evaluate this, we assessed models on Molecule Understanding, which demands the precise recognition of functional groups (critical for reactivity) and ring systems (the stable building blocks of drug scaffolds). We quantified performance using Mean Absolute Error (MAE) for counting tasks and Tanimoto similarity for structural scaffold extraction. As shown in Table~\ref{tab:molund_edit}, commercial general-purpose LLMs often suffer from "chemical hallucination"—misinterpreting the linearized SMILES syntax and failing to map 1D strings to 2D topologies. In contrast, \modelname{} achieves near-perfect precision. For example, in the Function-Group-Detection case (Table~\ref{tab:molund_edit}, bottom-left), where the model must isolate and count amino groups in a complex polycyclic architecture, \modelname{} achieves an MAE of 0.03, significantly outperforming standard instruction models like Qwen2.5-32B (MAE 0.36). This perceptual acuity extends to hierarchical structural reasoning; in the Ring-System-Detection task (Table 1, bottom-middle), which requires determining the existence of specific fused rings, our model attains 100\% accuracy compared to 87.5\% for Gemini-2.5-Pro~\cite{comanici2025gemini25}.

Crucially, this understanding translates into actionable Molecule Editing capabilities—operations (Add, Delete, Substitute) that serve as the fundamental "arithmetic" of chemical design. In the Molecule-Edit-Substitute visualization (Table~\ref{tab:molund_edit}, bottom-right), the challenge involves precisely targeting a hydroxyl group for removal and attaching a carboxyl group at the exact locus without disrupting the surrounding scaffold. \modelname{} leverages its sandbox tools to validate valency and connectivity, achieving a 95.0\% success rate in deletion tasks and a 97\% SMILES equivalence score. This confirms that \modelname{} has transcended probabilistic token prediction, effectively grounding its reasoning in verifiable chemical rules to perform rigorous structural modifications.

\begin{figure*}[!htp]
    \includegraphics[width=1.0\linewidth]{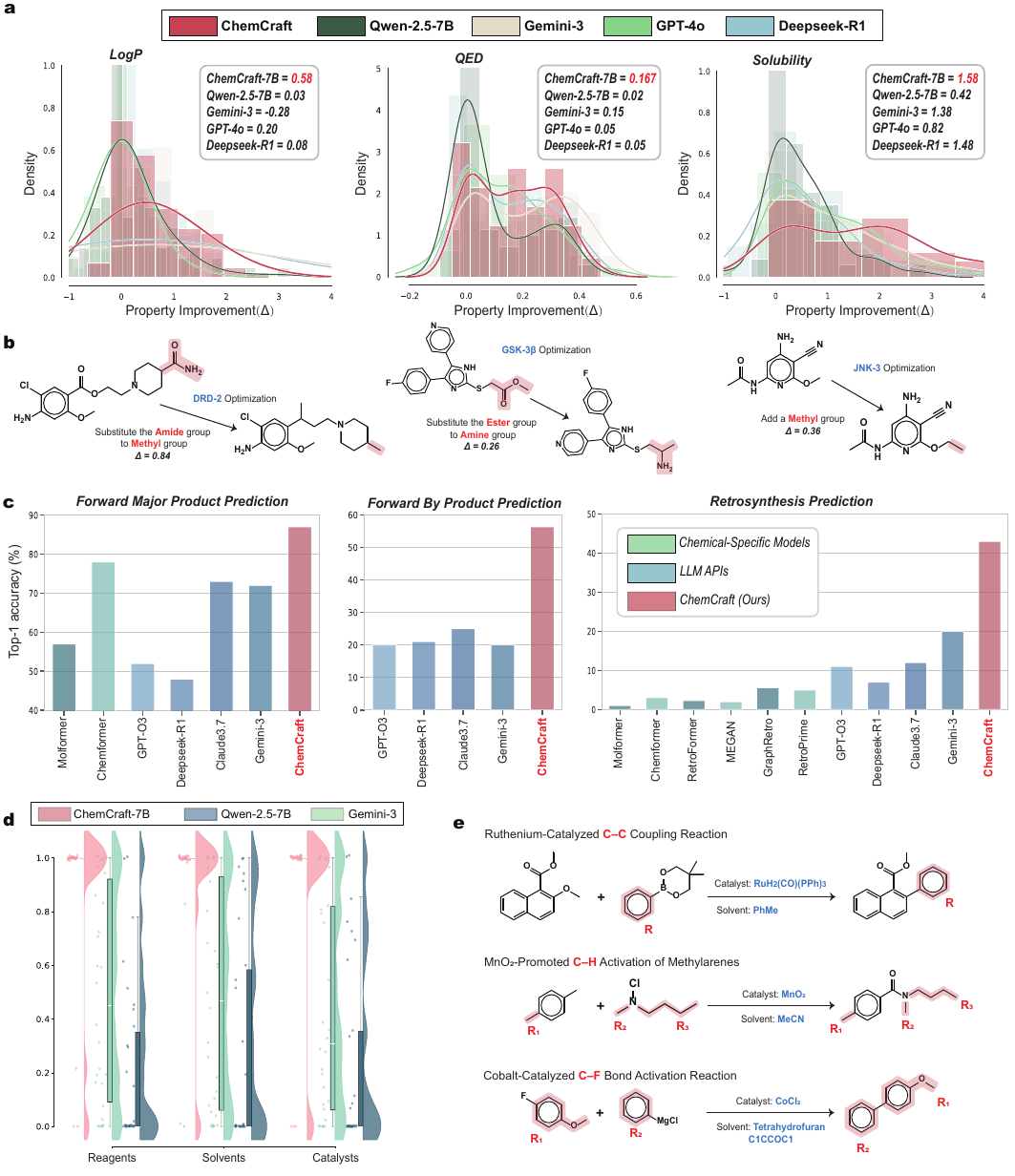}
    \caption{
    \textbf{Experimental Analysis for Molecular Optimization and Reaction Predictions}. \textbf{a} Distribution analysis for property improvements on LogP, QED, Solubility. \textbf{b} Optimized molecule visualizations for protein-activated properties, including DRD2, JNK3, GSK-3$\beta$. \textbf{c} Comparison between \modelname{}, chemical-specific models, and LLM-APIs on forward major / by product prediction and retrosynthesis prediciton. \textbf{d} Similarity distribution for reaction condition recommendations. \textbf{e} Reaction prediction visualizations on complex reaction types.
    } 
    \label{fig:model_validation}
\end{figure*}

\paragraph{Navigating the Optimization Landscape.}
Moving beyond static editing, we challenged the models with the open-ended task of Molecule Optimization—a proxy for de novo drug design where the goal is to modify a lead compound to maximize specific properties while retaining structural validity. We evaluated performance across two tiers of difficulty: (1) Physicochemical Properties (LogP, QED, Solubility), which follow defined heuristic rules, and (2) Protein-Ligand Binding Affinities (DRD2, GSK3-$\beta$, JNK3), which require inferring complex bio-interactions. Performance was quantified by the Property Improvement score ($\Delta$), measuring the net gain over the starting molecule.

As illustrated in the density plots of Figure~\ref{fig:model_validation}a, general LLMs (e.g., Qwen-2.5-7B~\cite{qwen2024qwen25}, Deepseek-R1) often struggle to escape local minima, producing distributions centered near zero ($\Delta \approx 0$). In contrast, \modelname{} (red distribution) consistently shifts the optimization trajectory toward positive gains. For instance, in Solubility optimization, \modelname{} achieves a mean improvement of $\Delta=1.58$, nearly quadrupling the performance of the baseline Qwen model ($\Delta=0.42$) and surpassing commercial reasoning models like Gemini-2.5-Pro ($\Delta=1.38$).

This statistical success translates into chemically meaningful design strategies, as visualized in the drug-target case studies in Figure~\ref{fig:model_validation}b. Here, the model effectively deploys the modular "Add/Delete/Substitute" operations validated in our Editing tasks to solve complex biological constraints. In the GSK-3$\beta$ case (middle), the model strategically substitutes an ester group with an amine ($\Delta=0.26$). This is a non-trivial bioisosteric replacement that significantly alters the hydrogen bond donor/acceptor profile to fit the binding pocket, demonstrating that the model is not merely randomizing atoms but applying medicinal chemistry logic. In the DRD-2 case (left), it executes a precise scaffold-hopping maneuver, replacing a flexible amide linker with a methyl group ($\Delta=0.84$) to potentially lock the conformation.
Crucially, these optimizations are achieved with high sample efficiency. Unlike traditional genetic algorithms that may require thousands of oracle calls (property evaluations), \modelname{} reaches these high-value candidates within minimal interaction steps. This efficiency suggests that our agent does not rely on brute-force search, but rather enables a "Hypothesis-Action-Verification" loop that is practical for real-world campaigns where wet-lab feedback is costly.

\paragraph{Predicting Complex Reaction Outcomes.} 
To evaluate chemical intuition at the systemic level, we assessed models on three pillars of reaction prediction: (1) Forward Prediction, encompassing both Major Products and, crucially, By-Products (vital for assessing impurity profiles and purification risks); (2) Retrosynthesis (w/o reaction type), testing the ability to deconstruct targets into available precursors; and (3) Condition Recommendation, requiring precise suggestions for reagents, solvents, and catalysts. We benchmarked against domain-specialized models (e.g., Chemformer~\cite{irwin2022chemformer}, GraphRetro~\cite{somnath2021learning}, RetroPrime~\cite{wang2021retroprime}) and general reasoning LLMs (e.g., GPT-o3, DeepSeek-R1~\cite{guo2025deepseek}).

As shown in Figure~\ref{fig:model_validation}c, while specialized models like Chemformer achieve competitive accuracy on Major Product prediction due to extensive pre-training on USPTO data, they falter significantly in By-Product Prediction and Retrosynthesis on out-of-distribution samples. We attribute this to architectural rigidity: sequence-based and graph-based expert models are typically optimized for single-outcome generation and often rely on precise atom-mapping, a dependency that limits their generalization to noisy or unmapped real-world inputs. In contrast, \modelname{} achieves a top-1 accuracy of over 50\% in by-product prediction and dominates the retrosynthesis task (40\% + vs < 20\% for baselines). Furthermore, in Condition Recommendation (Figure~\ref{fig:model_validation}d), the violin plots reveal that while general LLMs suffer from recommending convincing reaction conditions (distributions centered near 0.0), our \modelname{} produces highly concentrated predictions aligned with ground truth, particularly for complex catalysts.

The model’s superior grasp of reactivity is further evidenced in the case studies (Figure~\ref{fig:model_validation}e). We selected challenging modern reactions involving non-polar bond activation (C-C, C-H, C-F), transformations that are underrepresented in standard patent data (USPTO) and typically confound expert models. In the Ruthenium-catalyzed C-C coupling case~\cite{CCBond2021} (top), Gemini-2.5-Pro misses the competitive advantage of C-O activation over remote C-H activation. In the MnO$_2$ oxidation case~\cite{knochel2013amide} (middle), it underestimates the oxidation depth, predicting an amide rather than the correct carbonyl product. In the Cobalt-catalyzed C-F activation~\cite{wei2017cobalt} (bottom), the baseline fails to recognize that the specific additives are designed to suppress the benzyne pathway, leading to a regio-isomer error. \modelname{} correctly predicts these outcomes not by mere intuition, but by evidence-based reasoning: its sandbox retrieves analogous reaction templates from the library, allowing the agent to ground its mechanism in empirical precedent rather than hallucinating plausible but incorrect pathways.

\begin{figure*}[ht]
    \includegraphics[width=1.0\linewidth]{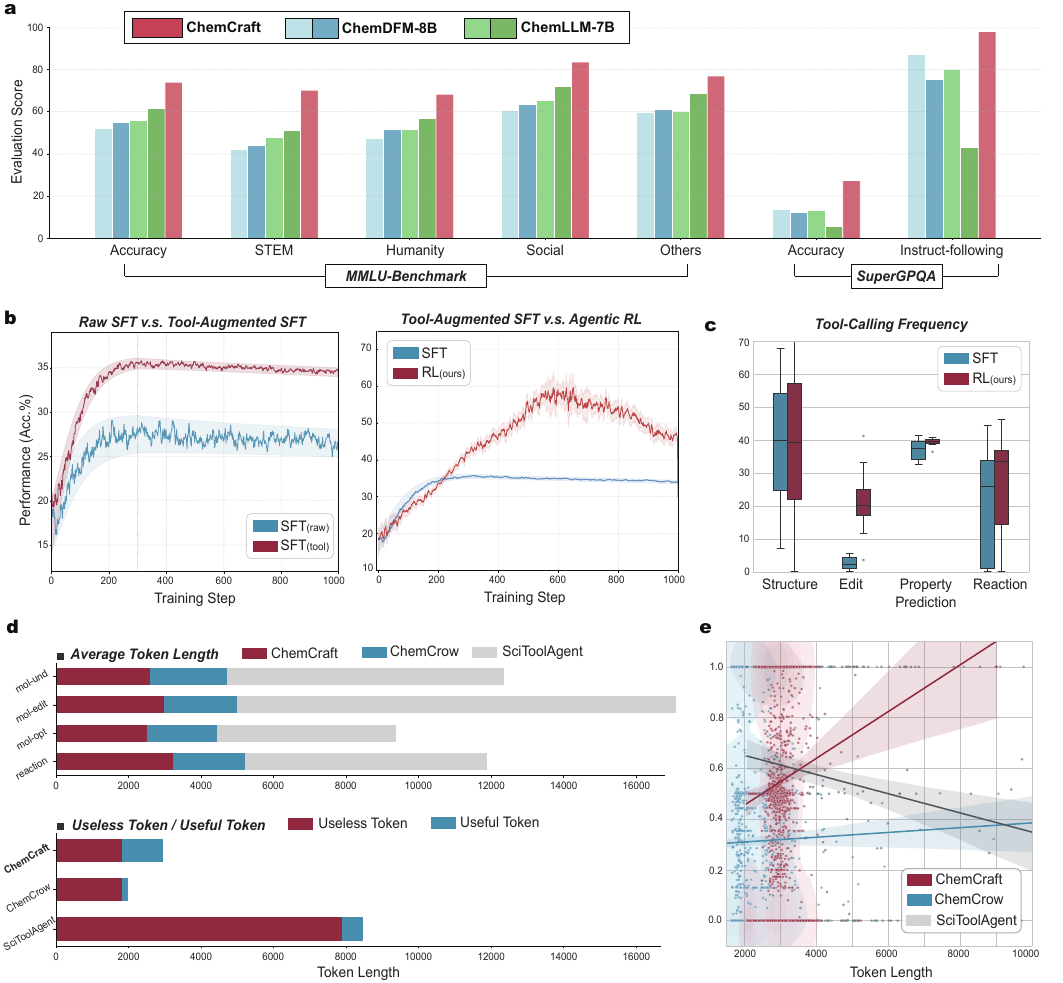}
    \caption{
    \textbf{a} Performance on general reasoning benchmarks. Our \modelname{} outperforms other chemical language models on different domains. \textbf{b} Training curves analysis for non-tool-calling SFT, tool-augmented SFT, and Agentic-RL. \textbf{c} Tool-calling frequency analysis between the SFT training paradigm and agentic-RL paradigm. \textbf{d} \modelname{} has low inference cost and a more useful token length compared to other chemical multi-agent systems. \textbf{e} Performance and token-length correlation analysis between \modelname{} and other chemical language models. 
    }
    \label{fig:model_sft_rl}
\end{figure*}

\subsection{Algorithmic Validation: Efficiency, Generalization, and the Power of Agentic RL} We posit that the superior performance of \modelname{} transcends mere domain data exposure, stemming instead from a cognitive decoupling architecture refined via reinforcement learning. To rigorously verify this hypothesis, we conducted a comprehensive ablation analysis evaluating training dynamics, generalization capabilities, and inference efficiency, as summarized in Figure~\ref{fig:model_sft_rl}.

\paragraph{The Necessity of Agentic Reinforcement Learning.} The limitations of standard SFT become evident when analyzing the training trajectories in Figure~\ref{fig:model_sft_rl}b. Models trained via "Raw SFT" without tool integration plateau early, as parameters alone are insufficient for precise scientific computation. While introducing tool-use trajectories ("Tool-Augmented SFT") resolves basic syntax errors and boosts performance, these models effectively learn the \textit{format} of tool usage rather than the \textit{strategy}. This is where the necessity of our agentic RL paradigm becomes undeniable. As illustrated in the right panel of Figure~\ref{fig:model_sft_rl}b, while the SFT baseline stagnates, the RL-driven training curve continues to climb, propelling the model toward a higher performance ceiling. This gain is mechanistically explained by Figure~\ref{fig:model_sft_rl}c: the RL-trained model exhibits a statistically significant increase in tool-calling frequency for complex subtasks like "Structure Analysis" and "Reaction Prediction." Unlike the SFT model, which remains a passive tool-user, the RL agent actively queries the sandbox to verify hypotheses, demonstrating a "Self-Correction" behavior where suboptimal initial actions are autonomously refined based on feedback.

\paragraph{Defying Catastrophic Forgetting.}
A pervasive risk in training specialized scientific models is the degradation of general reasoning capabilities, known as catastrophic forgetting. Figure~\ref{fig:model_sft_rl}a demonstrates that \modelname{} effectively circumvents this trade-off. While domain-adapted models like ChemLLM often compromise general capabilities due to rigid knowledge injection, \modelname{} preserves robust performance across general benchmarks. It achieves strong results on MMLU (STEM, Humanities, and Social Sciences) and SuperGPQA, significantly outperforming baselines. We attribute this robustness to the nature of our agentic training paradigm: rather than forcing the model to memorize static chemical knowledge or specific response templates via standard SFT, our approach instills a universal problem-solving methodology—analyzing the query, formulating a step-by-step plan, and orchestrating appropriate tools. This ensures that the model preserves its intrinsic cognitive structures for logic and analysis, allowing specialized chemical proficiency to coexist with, rather than overwrite, general intelligence.

\paragraph{The Efficiency of a Unified Agent Architecture.}
Finally, we address the computational feasibility of AI chemists. Existing Multi-Agent Systems (e.g., ChemCrow~\cite{bran2023chemcrow}, SciToolAgent~\cite{ding2025scitoolagent}) rely on massive commercial models and extensive inter-agent dialogue (e.g., hand-offs between a "Planner," "Critic," and "Executor"), resulting in prohibitive token costs. \modelname{} internalizes these roles into a single, streamlined reasoning stream. The impact of this architectural shift is quantified in Figure~\ref{fig:model_sft_rl}d: \modelname{} reduces the average token length by approximately 65\% compared to SciToolAgent while maintaining a higher density of "Useful Tokens". Figure~\ref{fig:model_sft_rl}e presents a normalized correlation analysis, where task metrics are mapped to a $[0, 1]$ scale to visualize the return on inference compute. The trend line illustrates an ideal reasoning efficiency, where increased token expenditure~(longer CoT) linearly translates to performance gains. \modelname{} closely adheres to this trajectory, demonstrating effective scaling of test-time compute. In contrast, multi-agent frameworks like SciToolAgent and ChemCrow diverge significantly, suffering from excessive token overhead with diminishing performance returns. This places \modelname{} in the optimal quadrant of high performance and low inference cost. The efficiency breakthrough resolves the data privacy and latency bottlenecks inherent to cloud-based solutions, making it feasible to deploy expert-level AI chemists on local hardware.

\section{Methods}
We present a unified framework designed to empower small language models~\cite{liu2025teaser,zhu2025learning,li2023textvqa}with autonomous reasoning and operational capabilities in the chemical domain. As illustrated in Fig.\ref{fig:framework}, our methodology is orchestrated across three interconnected pillars. First, to overcome the limitations of static knowledge storage, we construct a Chemical Agent Sandbox, an interactive environment that externalizes domain expertise through computational~\cite{landrum2013rdkit}, deep learning, and retrieval-based~\cite{li2023FreestyleRet,wu2026towards} tools. Second, to bridge the gap between abstract chemical knowledge and actionable problem-solving, we introduce a pipeline for constructing High-Quality Reasoning Trajectories, incorporating tool-integrated narratives and reflective refinement. Finally, we implement a Two-Stage Training Paradigm, which progressively transitions the model from supervised behavioral cloning to robust reinforcement learning, optimizing for scientific validity and rigorous chemical constraints via the SMILES-GRPO mechanism.

\subsection{Construction of the Chemical Agent Sandbox}
To endow the language model with comprehensive domain expertise, we established a Chemical Agent Sandbox—an interactive environment designed to decouple chemical reasoning from static knowledge storage. This framework covers the full spectrum of the chemical discovery pipeline, ranging from molecular structure analysis and de novo design to property optimization and retrosynthetic planning. To achieve robust performance across these diverse tasks, we encapsulate domain-specific tools into three distinct categories of agents:

\textbf{Computational Software Agents:} Leveraging robust cheminformatics libraries, primarily RDKit and RDChiral, these agents provide deterministic capabilities for molecular graph manipulation. They are responsible for verifying topological validity, analyzing stereochemistry, and executing precise structural editing, ensuring that the model’s generated molecules adhere to strict chemical rules.

\textbf{Deep Learning-based Agents:} Serving as the predictive intuition of the system, these agents interface with pre-trained models and frameworks like PyTDC. By providing rapid quantitative feedback on physicochemical properties (e.g., QED, LogP) and ADMET profiles, they guide the language model in navigating the high-dimensional chemical space during multi-objective optimization tasks.

\textbf{Retrieval-based Agents:} To anchor synthesis planning in empirical reality, these agents maintain a curated repository of reaction equations and templates. Through substructure matching and similarity search, they enable the model to retrieve analogous reaction pathways and validate reaction conditions, promoting mechanistic feasibility in synthesis prediction.

\subsection{Construction of High-Quality Reasoning Trajectory}
High-quality trajectory construction is pivotal for enabling the model to interact effectively with the chemical sandbox and solve complex domain tasks. We approach this construction in two phases: first by structuring the abstract chemical space into traceable reasoning steps, and subsequently by integrating dynamic tool interactions into these narratives.

\textbf{Generating Chemical Reasoning Trajectories}. To transcend simple factual recall, we align our data construction with ChemCoTBench, a framework that decomposes the chemical discovery process into 9 major tasks and 22 subtasks. This structure bridges fundamental capabilities, such as Molecule Understanding and Editing—with high, which stakes downstream applications like Molecule Optimization (e.g., binding affinity improvement) and Reaction Prediction (e.g., retrosynthesis planning). By breaking down these complex challenges into explicit sequences of modular chemical operations (e.g., functional group addition or substitution), we convert abstract chemical problems into actionable, step-by-step reasoning scaffolds. This rigorous decomposition ensures the model learns the operational logic required for real-world discovery rather than mere property memorization.

\textbf{Generating Tool-Integrated Reasoning Trajectories}. To capture the interactive nature of modern scientific workflows, we propose a Tool-Integrated Reasoning construction method. First, we leverage the Chemical Agent Sandbox to decouple reasoning from calculation, allowing the LLM to offload error-prone computations (e.g., RDKit parsing, QED evaluation) to external microservices. Second, to address the issue of disjointed API logs, we implement a "Reflective Refinement" mechanism. Instead of retaining mechanical "Action-Observation" pairs, this process injects verified tool outputs back into the context and prompts a teacher model to rewrite the reasoning trace. This transforms rigid tool logs into fluid, expert-level scientific narratives, where the agent interprets evidence, validates hypotheses, and dynamically adjusts its strategy, mirroring the cognitive process of a professional chemist.

\subsection{Model Training: A Two-Stage Paradigm}

To effectively instill both chemical domain knowledge and agentic reasoning capabilities into the model, we employ a progressive two-stage training paradigm. This approach first establishes a stable behavioral policy through supervised learning and subsequently refines the model's problem-solving strategies using reinforcement learning with chemistry-specific feedback.

\subsubsection{Stage 1: Cold-Start Supervised Fine-Tuning}

The primary objective of the ``Cold-Start'' phase is to initialize the model's understanding of chemical syntax and establish the fundamental ``Think $\rightarrow$ Call Tool $\rightarrow$ Observe'' behavioral pattern. We utilize the synthesized tool-integrated trajectories to perform Supervised Fine-Tuning (SFT) on the base model (e.g., Qwen-2.5/3).

We treat the agentic interaction as a sequential generation task. Given an input prompt $x$ and a target trajectory $y = (y_1, y_2, \dots, y_T)$ consisting of interleaved reasoning thoughts, tool calls, and observations, we optimize the model parameters $\theta$ by minimizing the negative log-likelihood of the next token. The loss function $\mathcal{L}_{\text{SFT}}$ is defined as:
\begin{equation}
\mathcal{L}_{\text{SFT}}(\theta) = - \mathbb{E}_{(x, y) \sim \mathcal{D}_{\text{SFT}}} \left[ \sum_{t=1}^{T} \log P_\theta(y_t \mid x, y_{<t}) \right]
\label{eq:sft_loss}
\end{equation}

where $\mathcal{D}_{\text{SFT}}$ represents the curated dataset of tool-integrated trajectories, and $y_{<t}$ denotes the history of tokens preceding step $t$. This phase ensures that the model learns to strictly adhere to the formatting requirements of the \textit{Chemical Agent Sandbox} and mimics the expert-level reasoning logic embedded in the training data, providing a robust policy initialization $\pi_{\text{ref}}$ for the subsequent reinforcement learning stage.

\subsubsection{Stage 2: Reinforcement Learning with SMILES-GRPO}

While SFT provides a strong foundation, it is limited by the static nature of imitation learning. To transcend mere mimicry and enable the model to explore novel chemical spaces, we advance to a reinforcement learning framework. We adopt Group Relative Policy Optimization~(GRPO), which eliminates the need for a value function approximation by normalizing advantages within a group of sampled outputs.

\paragraph{The GRPO Objective.}
For each chemical query $q$, we sample a group of $G$ outputs $\{o_1, o_2, \dots, o_G\}$ from the old policy $\pi_{\theta_{\text{old}}}$. The optimization objective maximizes the surrogate objective while constraining the policy update via KL-divergence. The GRPO loss function is formulated as:
\begin{equation}
\begin{aligned}
\mathcal{J}_{\text{GRPO}}(\theta) = \mathbb{E}_{q \sim \mathcal{D}, \{o_i\}_{i=1}^G \sim \pi_{\theta_{\text{old}}}} \Bigg[ \frac{1}{G} \sum_{i=1}^G \Bigg( &\min \left( r_i \frac{\pi_\theta(o_i|q)}{\pi_{\theta_{\text{old}}}(o_i|q)}, r_i \text{clip} \left( \frac{\pi_\theta(o_i|q)}{\pi_{\theta_{\text{old}}}(o_i|q)}, 1-\epsilon, 1+\epsilon \right) \right)- \beta \mathbb{D}_{\text{KL}}(\pi_\theta || \pi_{\text{ref}}) \Bigg) \Bigg]
\end{aligned}
\label{eq:grpo_loss}
\end{equation}

where $\epsilon$ is the clipping parameter, $\beta$ controls the KL-divergence penalty to prevent policy collapse, and $r_i$ is the advantage term, calculated by normalizing the rewards within the group: $r_i = \frac{R_i - \text{mean}(\{R_j\}_{j=1}^G)}{\text{std}(\{R_j\}_{j=1}^G)}$.

\paragraph{Multidimensional Chemical-Aware Reward.}
To rigorously evaluate the scientific validity of the reasoning chain, we engineered a dense chemical reward function $R_{\text{total}}$. Unlike generic scalar feedback, our reward signal integrates format compliance with deep chemical verification metrics. The total reward for a generated output $o$ is defined as:
\begin{equation}
R_{\text{total}}(o) = \lambda_{1} \cdot R_{\text{format}} + \lambda_{2} \cdot R_{\text{chem}}
\label{eq:total_reward}
\end{equation}

where $R_{\text{format}}$ is an indicator function for syntactic correctness (e.g., correct tool calling tokens). The chemical reward $R_{\text{chem}}$ is a weighted sum of four key components:
\begin{equation}
R_{\text{chem}} = w_{str} \cdot \text{Sim}_{\text{scaffold}} + w_{func} \cdot \text{Fidelity}_{\text{func}} + w_{opt} \cdot \Delta \text{Prop} + w_{rxn} \cdot \text{Valid}_{\text{rxn}}
\label{eq:chem_reward}
\end{equation}

Specifically, this composite reward mechanism first enforces structural integrity by measuring the Tanimoto similarity of Bemis-Murcko scaffolds, ensuring that generated molecules retain the desired core architecture. It simultaneously evaluates functional fidelity by explicitly penalizing the loss of critical functional groups mandated by the input prompt. Beyond structural constraints, the objective function quantifies the magnitude of property optimization, rewarding positive shifts in target metrics such as QED or binding affinity. Finally, to guarantee synthetic realizability, it incorporates a validity check that assesses whether the predicted synthesis pathways align with established chemical reaction templates.

By optimizing against these granular metrics, SMILES-GRPO drives the model to strategize its tool usage, ensuring that the generated molecules are not only textually valid but also scientifically viable and optimized for their target properties.
\bibliography{related_papers}

\end{document}